# Green vehicle routing problem that jointly optimizes delivery speed and routing based on the characteristics of electric vehicles

YY.Feng

**Abstract**

The abundance of materials and the development of the economy have led to the flourishing of the logistics industry, but have also caused certain pollution. The research on GVRP (Green vehicle routing problem) for planning vehicle routes during transportation to reduce pollution is also increasingly developing. Further exploration is needed on how to integrate these research findings with real vehicles. This paper establishes an energy consumption model using real electric vehicles, fully considering the physical characteristics of each component of the vehicle. To avoid the distortion of energy consumption models affecting the results of route planning. The energy consumption model also incorporates the effects of vehicle start/stop, speed, distance, and load on energy consumption. In addition, a load first speed optimization algorithm was proposed, which selects the most suitable speed between every two delivery points while planning the route. In order to further reduce energy consumption while meeting the time window. Finally, an improved Adaptive Genetic Algorithm is used to solve for the most energy-efficient route. The experiment shows that the results of using this speed optimization algorithm are generally more energy-efficient than those without using this algorithm. The average energy consumption of constant speed delivery at different speeds is 17.16% higher than that after speed optimization. Provided a method that is closer to reality and easier for logistics companies to use. It also enriches the GVRP model.

## 1. Introduction

In conventional green vehicle routing problems, the energy consumption of electric vehicles is mostly calculated using fixed formulas. And many conditions are idealized. First of all, the impact of the vehicle itself includes battery capacity, battery charging and discharging characteristics, motor efficiency and aerodynamic characteristics. For example, the discharge efficiency of the battery supplying energy is different at different discharge voltages. The mechanical efficiency of the motor is also different at different powers. External influencing factors include driving habits, ambient temperature, road conditions, driving speed and vehicle load. The development of electric vehicles is changing with each passing day, and various new technologies are emerging in an endless stream. The electrical performance of batteries and motors is also constantly improving. The energy consumption of vehicles of different models is very different under the same road conditions. When using a fixed energy consumption calculation formula to calculate the actual energy consumption of vehicles of different models, large errors may occur. The previous formula does not conform to the current car. As a result, the energy consumption of the vehicle calculated by the formula in actual application has a certain error with the actual one. The accumulated error will make the theoretical optimal route obtained not actually the optimal route. Therefore, modeling each model of vehicle separately can get more accurate results.

This paper first studies the use of simulation methods to model the energy consumption of a real model A electric vehicle. The simulation model can set detailed settings for vehicles and road conditions to obtain a more realistic model. The simulation results of the energy consumption model are compared with the results of actual tests to verify the feasibility of this method. Second, a load first speed optimization algorithm(LsA) is introduced into the GVRP model to actively change the delivery speed to reduce the vehicle's energy consumption as much as possible while meeting the time window requirements. Finally, an Adaptive Genetic Algorithm is used to solve the problem. Compare the energy consumption gap between constant speed delivery and active variable speed delivery. Because most of the electricity now comes from thermal power plants, lower energy consumption means a more environmentally friendly delivery method.

**Electric vehicle energy consumption model**

This chapter proposes a method for obtaining energy consumption formulas under set conditions. Firstly, Simulate and model industrial level energy consumption for real vehicles. Then verify the accuracy of the energy consumption model. The second step is to build simulation scenarios based on the set GVRP conditions. Simulate at different speeds and loads to obtain discrete data on energy consumption corresponding to driving distance. By performing multiple nonlinear regression on these data at different speeds, the specific energy consumption of the real vehicle Model A for any distance traveled with any load under the conditions set by GVRP can be obtained at the same speed. Without the need to be proficient in automotive engineering knowledge, more accurate vehicle energy consumption data can be obtained in actual distribution tasks in subsequent vehicle route planning, making the entire model more realistic and practical.

*1.1. Build the vehicle energy consumption module*

*1.1.1. Motor*

There are constant power and constant torque zones at different speeds. When outputting different torques at the same speed, the efficiency curve of the motor is also different. As shown in formula 2-1, $\eta_E$ is the efficiency of the motor, which depends on the characteristics of the motor itself. $P_{E-in}$ is the input power of the motor, and $P_{E-out}$ is the output power of the motor. In most VRP studies, the calculation of vehicle energy consumption is usually treated as a constant, which is inaccurate in practical applications. When vehicles require different speeds and torques, $\eta_E$ is not the same. To obtain a more accurate energy consumption model, fill in the corresponding $\eta_E$ in the motor module.

$$\eta_E = P_{E-in} / P_{E-out} \qquad (2-1)$$

*1.1.2. Battery*

Batteries have a significant impact on vehicle endurance. As the battery discharges, internal chemical reactions will cause the voltage to gradually decrease. Therefore, the output voltage of the battery cannot be directly regarded as a constant and needs to be adjusted in real time according to different levels of power.

*1.2. Model accuracy verification*

In order to verify its accuracy, it is necessary to compare the simulation results with the actual results. This article uses the NEDC test results for comparison. The NEDC simulation test was conducted in avl cruise, and the test results are shown in Figure 2.1. The value of $R_{NEDC}$, which reflects the economy of the car, is 4.6253km/kWh. According to formula 2-2, the NEDC simulation test range of Model A, $L'_{NEDC}$, is 232.32km, and $C_A$ is the battery capacity of Model A. The difference $\Delta_{NEDC}$ between the simulation result and the measured result $L_{NEDC}$ is calculated by formula 2-3. The calculation results show that $\Delta_{NEDC}=1\%$. The accuracy of the simulation process is proved, and the energy consumption model is reliable.

$$L'_{NEDC} = R_{NEDC} \cdot C_A \qquad (2-2)$$

$$\Delta_{NEDC} = (L_{NEDC} - L'_{NEDC})/L_{NEDC} \cdot 100\% \qquad (2-3)$$

```
                             *** RESULTS ***

                                                                   |----------|
|-----------------|----------|------------|------------|----------| DISTANCE |
|     Cycle       |  Fault-  | Vehicle -  |Calculation |   Slip   |          |
|                 |   time   |   mass     |            |          |          |
|                 |   [s]    |   [kg]     |            |          |   [m]    |
|=================|==========|============|============|==========|==========|
|     NEDC        |  20.05   |   2115.0   | Simulatio* |   Off    | 10908.19 |
|-----------------|----------|------------|------------|----------|----------|

           Overall Fuel Consumption:            0.0000    [kg]
           Idle Fuel Consumption:               0.0000    [kg]
           Acceleration Fuel Consumption:       0.0000    [kg]
           Constant Drive Fuel Consumption:     0.0000    [kg]
           Deceleration Fuel Consumption:   |   0.0000    [kg]
           Overall Energy Consumption:          2.3584    [kWh]
           Electric Fuel Economy:               4.6253    [km/kWh]
           Electric Fuel Economy:               0.2162    [kWh/km]
```

Figure 2.1 NEDC simulation test results of Model A

*1.3. GVRP simulation condition setting*

*1.3.1. Cycle run profile setting*

Most of the vehicle energy consumption models used in VRP research do not take into account the process of vehicle starting and stopping. Because the start stop process involves the conversion of static friction and rolling friction, the calculation process is relatively complex. However, starting and stopping have a significant impact on the energy consumption of automobiles. This paper sets the car to start and stop at an acceleration of $2m/s^2$ every 5 kilometers, and cycles until the vehicle reaches its minimum battery capacity. In practice, the distribution roads can be surveyed and appropriate acceleration can be set according to the statistical results.

*1.3.2. Speed setting*

This paper sets the delivery speed to three levels: 40, 50, and 60 km/h. In actual driving, it is difficult for vehicles to maintain a unique speed, which represents the average speed between any two delivery points. Setting fewer speed gears can save solving time[1]. Excessive grading will only increase computational complexity and will not significantly improve the solution results.

*1.4. Calculation and data processing*

At each speed, the vehicle is simulated by adding equal loads until it is fully loaded. The data relationship between the driving distance and energy consumption at each speed and different loads is obtained. The results output by the software are shown in Figure 2.2, Figure 2.3. The lines in the figure represent the load reduction from left to right.

Use multiple nonlinear regression to process these data. Obtain the accurate energy consumption calculation formula 2-4 for Model A at the same speed. $C_v$ is the energy consumption at speed $v$, $p$ is the coefficient of the formula, depending on different speeds, $L$ is the distance traveled by the vehicle, and W is the load on the vehicle. Confidence interval of 95%. The comparison between the results calculated by the energy consumption calculation formula and the simulation data is shown in Figure 2.3 . It can be seen from the figure that the result obtained by the energy consumption calculation formula has a high degree of coincidence with the simulation data. This shows that at the same speed, the energy consumption of any load and any distance can be obtained through the energy consumption calculation formula.

$$C_v = p00 + p10 \cdot L + p01 \cdot W + p20 \cdot L^2 + p11 \cdot L \cdot W \qquad (2-4)$$

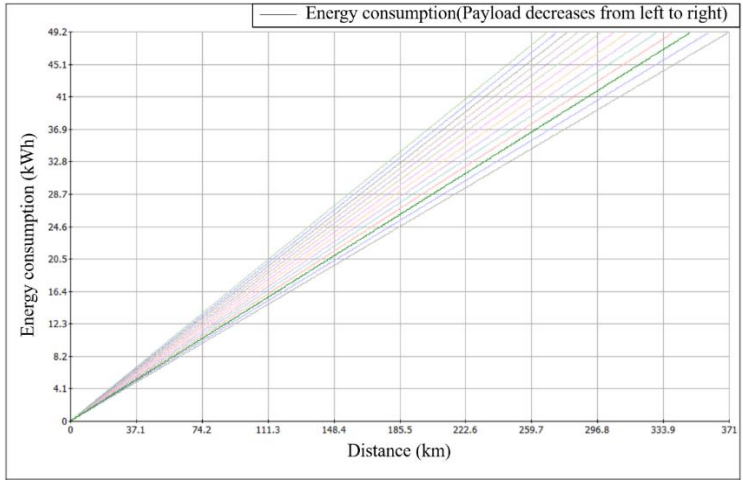

Figure 2.2 The relationship between distance and energy consumption under different load at 40km/h

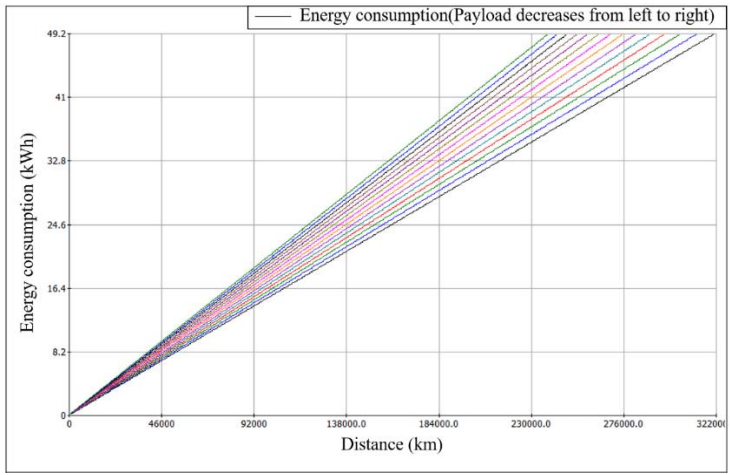

Figure 2.3 The relationship between distance and energy consumption under different load at 50km/h

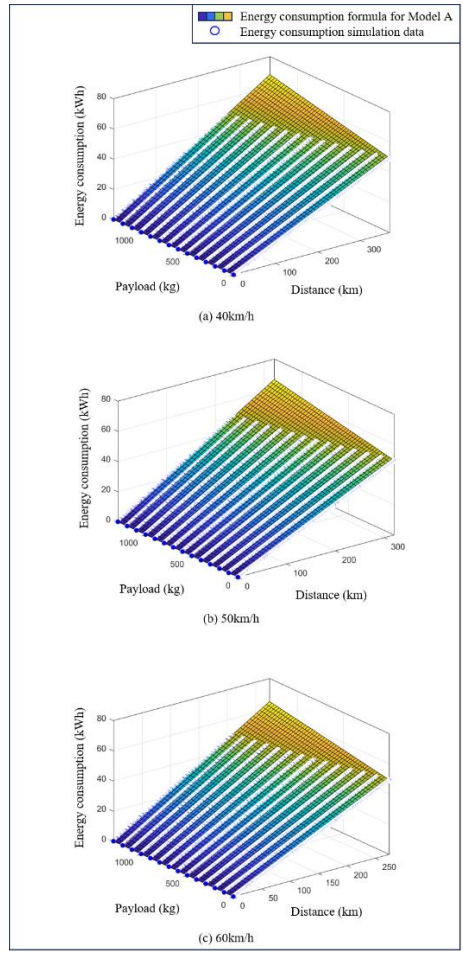

Figure 2.3 Comparison of energy consumption calculation formulas with simulation data

**GVRP model construction**

    This chapter mainly discusses the GVRPtw model based on vehicle characteristics. The minimum energy consumption is taken as the objective function of the GVRP model. Assuming there are n customers (including a starting point and an ending point), the number of vehicles is m, each customer $i$ has a demand quantity $d_i$, and the capacity of each vehicle is W, $c_{ij}^v$ is the electricity consumption from customer $i$ to customer $j$.

*1.5. Problem description*

    (1) The vehicle starts from the distribution center and only carries out the transportation of goods, and no longer receives goods on the way.
    (2) A distribution center from which vehicles depart and eventually return to the distribution center.
    (3) All delivery vehicles are of the same model.
    (4) The location of the distribution center and the customer is determined.
    (5) The vehicle is fully charged before departure.
    (6) The demand of each customer point is also determined.

(7) A customer can only be served by one vehicle.
(8) Do not consider the impact of unloading operation on energy consumption.
(9) Each customer point must and can only be accessed once.
(10) Ensure that the total endurance is greater than the total vehicle driving route.
(11) The weight and volume of transportation shall not exceed the vehicle limit.
(12) Can be delivered in advance, but not more than $T_a$ minutes in the time window.
(13) The speed between each two distribution points can only be selected from 40,50,60 km/h.
(14) The time of discharge shall be taken into account.

## 1.6. Decision variable

$$x_{ij}^k = \begin{cases} 1, & \text{If vehicle } k \text{ is traveling directly from customer } i \text{ to } j \\ 0, & \text{others} \end{cases} \quad (3-1)$$

$$y_{ij}^q = \begin{cases} 1, & \text{If the vehicle chooses speed } v^q \text{ on the route } (i,j) \\ 0, & \text{others} \end{cases} \quad (3-2)$$

## 1.7. Mathematical model

$$\min C = \min \sum \sum_{i,j \in P, i \neq j} c_{ij}^k \quad (3-2)$$

s. t.

$$\sum_{i \in CL, i \neq j} x_{ij}^k = 1 \quad \forall j \in CL, \forall k \in K \quad (3-3)$$

$$\sum_{j \in CL, j \neq 0} x_{0j}^k \leq n \quad \forall k \in K \quad (3-4)$$

$$0 \leq w_{0j} \leq W \quad j \in P \quad (3-5)$$

$$\sum_{j \in P} w_{ji} - \sum_{j \in P} w_{ij} \leq d_i \quad \forall i \in CL \quad (3-6)$$

$$\theta C_A \leq b_i \leq C_A \quad \forall i \in P \quad (3-7)$$

$$\sum_{j=1}^{n} x_{0j}^k = 1 \quad \forall k = 1, \dots, m \quad (3-8)$$

$$\sum_{i=1}^{n} x_{i0}^k = 1 \quad \forall k = 1, \dots, m \quad (3-9)$$

$$y_{ij}^{40} + y_{ij}^{50} + y_{ij}^{60} = 1 \quad \forall i, j \in P \quad (3-10)$$

$$y_{ij}^{40} + y_{ij}^{50} + y_{ij}^{60} \leq x_{ij} \quad \forall i, j \in P \quad (3-11)$$

$$k_{i-1} + h_{i-1} \leq TW_i' + T_a \quad i \in CL \setminus \{1\} \quad (3-12)$$

Formula 3-2 is the objective function, C represents the energy consumption, and the objective function is to minimize the energy consumption. Formula 3-3 indicates that each customer has only one visit. Formula (3-4) indicates that the number of vehicles for delivery cannot be more than the number of customers, and there cannot be empty vehicles. Formula 3-5 indicates that the vehicle does not exceed the vehicle load limit when it departs. Formula 3-6 controls whether the cargo demand of each distribution point is met. Formula 3-7 indicates the range of energy usage in vehicle distribution. Formulas 3-8 and 3-9 indicate that the route of each vehicle must start and end at the distribution center. 3-10 indicates that on each route (*i, j*), the speed can only be selected from one of the three speeds. 3-11 indicates that the speed needs to be selected only when the route (*i, j*) is selected ($x_{ij}=1$). 3-12 indicates the time window restriction. The above is a rough mathematical model of GVRP.

**Adaptive Genetic Algorithm embedded with LsA**

This chapter proposes the load first speed optimization algorithm (LsA). Being able to save as much energy as possible while meeting the time window. This algorithm is embedded into Adaptive Genetic Algorithm(AGA) to solve the GVRP model in the previous chapter.

*1.8. Algorithm Overview*

The general Genetic Algorithm code[2] is improved by introducing the nearest neighborhood and adaptive parameters, and the LsA is embedded in it. The process is shown in the Figure 4.1. In the Genetic Algorithm, the setting of parameters has a significant impact on the performance of the algorithm. Different parameters affect the efficiency of the search, the quality of the solution, and the speed of algorithm convergence.

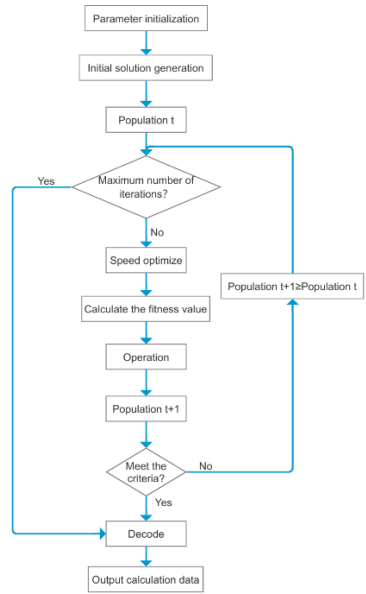

Figure 4.1 Adaptive Genetic Algorithm process with speed optimization algorithm

*1.9. Population Initialization*

The quality of the initial solution has a great impact on the solution effect of the entire algorithm. A good initial solution can accelerate the convergence of the objective function in the early stage. But at the same time, it is also necessary to ensure the diversity of the entire population to avoid falling into the local optimal solution. Therefore, two methods are used to generate the initial population. The first is the Nearest Neighbor Heuristic, which is fast and effective. Each time, the nearest unvisited distribution point is selected as the next distribution point to be visited. The second is a randomly generated initial population, which can effectively improve the diversity of the entire population. While generating the initial solution, it is ensured that the initial solution meets the constraints.

*1.10. LsA*

Once the population is obtained, speed optimization based on load first can be performed on each individual. This algorithm is embedded in the Genetic Algorithm, and the speed between any two points can be obtained after planning the route.

*1.11. Fitness Function*

The distance between every two points is fixed. By planning the route of each individual vehicle, the load of the vehicle from *i* to *j* can be obtained. And the speed plan that has already been obtained. Calculate the energy consumption between each two points according to formula 2-4. Based on the next delivery data, the equivalent distance will be obtained by reversing the energy consumption, and then the equivalent distance will be added to the new delivery distance to obtain the energy consumption of the next point. Then calculate the fitness value of each individual. In the subsequent iteration process, some individuals that do not meet the constraints will be generated. Screen out these individuals and increase their maladaptive values through a penalty function. Let it be eliminated in subsequent iterations.

*1.12. Genetic Operators*

Genetic operators imitate the evolutionary process of genes and carry out specific operations on individuals. In this paper, four operators of selection, crossover, mutation and reversal are used.

When solving complex combinatorial optimization problems, Tournament Selection can usually obtain better optimization results quickly. Because the Tournament Selection selects the winner by comparing the fitness values of multiple individuals, it helps to quickly screen out the individuals with higher fitness, thus accelerating the evolution process of the population and improving the convergence speed of the algorithm.

The Crossover operator pairs the chromosomes of two parent individuals. Then compare the crossover probability $P_c$ with the size of the randomly generated number. If the probability of Crossover is high, Partially Matched Crossover (PMX) is used to exchange some genes between them. Combine the excellent genes of different individuals to produce offspring that are better than their parents.

The reversal operator reverses some individuals to test whether the result is better.

**Result and Discussion**

In this experiment, the computer CPU used is Intel 12th i7, with 16GB of memory and operating system Windows 11. The dataset used is Solomon[3], a classic dataset in VRP research. Each instance is run 10 times to take an average as a result. This chapter will include the following parts: speed strategy experiment, algorithm applicability verification, vehicle matching analysis.

*1.13. Speed strategy experiment*

The speed between every two delivery points will be planned using the LsA. The planned speed is selected between 40, 50, and 60km/h. The vehicle energy consumption is calculated according to formula 2-4. Second, the energy consumption of constant speed distribution is calculated. The constant delivery speeds are 40, 50, and 60km/h respectively. Select the rc200 series in the Solomon, 100 delivery points. The energy consumption of different speed strategies is shown in Table 5-1, which records the best results and average results of each test. The results of constant speed delivery are compared with those of LsA. The average value of the gap between each instance under the same speed strategy and the overall average value of the gap are also counted. The evolution process and route planning of the optimal solution of rc201 are shown in the Figure .

The data in the table shows that the energy consumption obtained by using LsA is lower than that of constant speed delivery on the whole. For fixed speed delivery, the faster the selected speed is, the larger the gap with the lsa. The difference in average energy consumption is as large as 40.54%. The difference in energy consumption between constant speed delivery and LsA is the smallest at 40km/h, with an optimal energy consumption difference of 2.30% and an average energy consumption difference of 0.90%. However, there are some examples where the energy consumption of 40km/h is lower than that of LSA.

Table 5-1 Optimization results under different speed strategies

| Strategy | Instance | rc201 | rc202 | rc203 | rc204 | rc205 | rc206 | rc207 | rc208 | Same strategy gap |
|---|---|---|---|---|---|---|---|---|---|---|
| LsA | Best | 165.22 | 155.39 | 150.53 | 130.92 | 148.75 | 158.19 | 152.46 | 141.19 | |
|  | Avg | 173.52 | 174.06 | 154.2 | 137.51 | 162.31 | 165.72 | 162.31 | 146.42 | |
| 40 | Best | 175.43 | 162.63 | 149.77 | 135.21 | 158.57 | 160.03 | 155.75 | 133.98 | |
|  | Δ/% | 6.18% | 4.66% | -0.50% | 3.28% | 6.60% | 1.16% | 2.16% | -5.11% | 2.30% |
|  | Avg | 184.7 | 169.48 | 155.92 | 136.54 | 166.05 | 167.14 | 163.24 | 145.34 | |
|  | Δ/% | 6.44% | -2.63% | 1.11% | -0.70% | 2.30% | 0.86% | 0.57% | -0.74% | 0.90% |
| 50 | Best | 192.76 | 195.31 | 162.71 | 158.5 | 192.46 | 190.53 | 163.68 | 161.96 | |
|  | Δ/% | 16.67% | 25.69% | 8.09% | 21.07% | 29.38% | 20.44% | 7.36% | 14.71% | 17.93% |
|  | Avg | 208.86 | 200.99 | 174.19 | 160.63 | 201.02 | 198.27 | 183.39 | 164.7 | |
|  | Δ/% | 20.37% | 15.47% | 12.96% | 16.82% | 23.85% | 19.64% | 12.99% | 12.48% | 16.82% |
| 60 | Best | 216.16 | 205.48 | 197.31 | 179.42 | 209.21 | 221.47 | 193.99 | 185.12 | |
|  | Δ/% | 30.83% | 32.24% | 31.08% | 37.05% | 40.65% | 40.00% | 27.24% | 31.11% | 33.77% |
|  | Avg | 237.08 | 225.04 | 203.82 | 184.24 | 218.30 | 232.90 | 210.06 | 195.55 | |
|  | Δ/% | 36.63% | 29.29% | 32.17% | 33.98% | 34.49% | 40.54% | 29.41% | 33.55% | 33.76% |

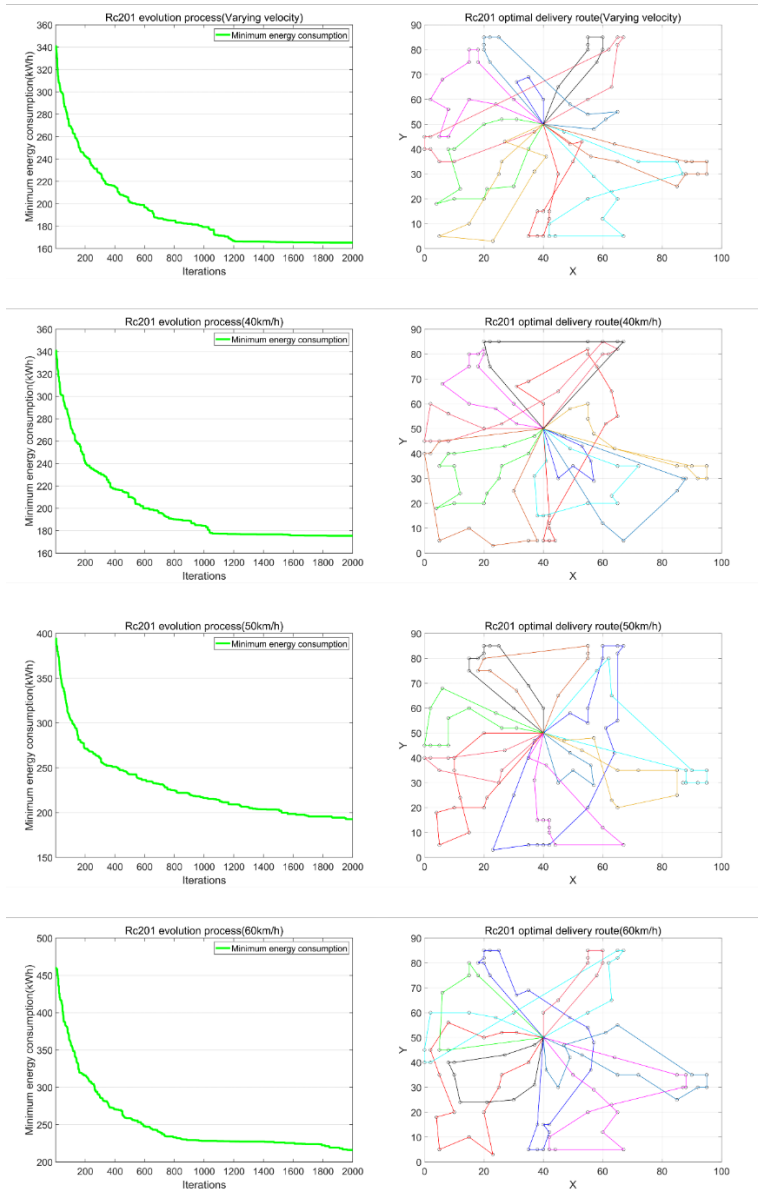

Figure 5.1 Evolution process and routing of rc201 under different speed strategies

## Conclusion

This paper mainly studies the GVRP of joint optimization route and speed. The energy consumption model of electric vehicles is derived from the simulation modeling of real vehicles. The energy consumption effects of vehicle start/stop, speed, distance and load are considered. The speed is planned using the load first speed optimization algorithm. Then the problem is solved using the Adaptive Genetic Algorithm. Experiments show that the results obtained by this method are better than those of constant speed delivery in most cases, and have good wide applicability.

Enriched the theoretical model of VRP. It also provides a feasible solution for some enterprises or cities that have restrictions on carbon emissions. It is convenient for logistics companies to plan more suitable routes according to the energy consumption characteristics of their vehicles. More physical factors are taken into account to reduce the distortion of the energy consumption model and avoid misleading route planning due to incorrect energy consumption calculation. Setting the speed for each segment will also be easier to achieve in the increasingly popular autonomous driving trucks in the future.

**References**

1   Xu, Z., Elomri, A., Pokharel, S. & Mutlu, F. A model for capacitated green vehicle routing problem with the time-varying vehicle speed and soft time windows. *Computers Industrial Engineering* **137**, 106011 (2019).
2   Lewis, F. *Solving-TSP-VRP*, <https://github.com/liukewia/Solving-TSP-VRP/tree/master/GA_VRPTW> (2023).
3   Instances of Solomon's VRPTW, <https://www.sintef.no/projectweb/top/vrptw/> (2008).